\documentclass[letterpaper]{article} 
\usepackage{aaai2026}  
\usepackage{times}  
\usepackage{helvet}  
\usepackage{courier}  
\usepackage[hyphens]{url}  
\usepackage{graphicx} 
\usepackage{natbib}  
\usepackage{caption} 
\usepackage{algorithm}
\usepackage{algorithmic}

\usepackage{booktabs}
\usepackage{multirow}
\usepackage{amsmath}
\usepackage{amsfonts}
\usepackage{newfloat}
\usepackage{listings}
\DeclareCaptionStyle{ruled}{labelfont=normalfont,labelsep=colon,strut=off} 
\floatstyle{ruled}
\newfloat{listing}{tb}{lst}{}
\floatname{listing}{Listing}
\nocopyright

\title{Advancing Wildlife Conservation through Multimodal Animal Re-Identification with Environmental Metadata}
\author{
    Yuzhuo Li\textsuperscript{\rm 1}\equalcontrib,
    Di Zhao\textsuperscript{\rm 1}\equalcontrib\thanks{Corresponding author.},
    Tingrui Qiao\textsuperscript{\rm 1},
    Yihao Wu\textsuperscript{\rm 1},
    Bo Pang\textsuperscript{\rm 1},
    Yun Sing Koh\textsuperscript{\rm 1}
}

\affiliations{
    \textsuperscript{\rm 1}School of Computer Science, University of Auckland\\
    Auckland, New Zealand\\
    \{yil708, tqia361, ywu840, bpan882\}@aucklanduni.ac.nz\\
    \{di.zhao, y.koh\}@auckland.ac.nz
}

\usepackage{bibentry}

\begin{document}

\maketitle

\begin{abstract}
Identifying individual animals is crucial for effective wildlife monitoring and conservation efforts.
Recent advancements in computer vision have shown promise in animal re-identification (Animal ReID) by leveraging data from camera traps.
However, existing Animal ReID datasets rely exclusively on visual data, overlooking environmental metadata that ecologists have identified as highly correlated with animal behavior and identity, such as temperature and circadian rhythms.
Meanwhile, modern vision–language models (VLMs) offer rich multimodal reasoning capabilities, but existing resources underutilize their text-processing potential. 
To address these limitations, we propose MetaWild, a multimodal Animal ReID dataset comprising 20,890 images across six species, paired with environmental metadata extracted from embedded camera trap overlays and scene contexts. 
Additionally, to facilitate the use of metadata in existing ReID methods, we propose the Meta-Feature Adapter (MFA), a lightweight module that can be incorporated into existing VLM-based ReID methods, allowing ReID models to leverage both environmental metadata and visual information to improve ReID performance.
Experiments on MetaWild show that combining baseline ReID models with MFA to incorporate metadata consistently improves performance compared to using visual information alone, validating the effectiveness of incorporating metadata in re-identification. 
\end{abstract}


\section{Introduction}
\begin{figure}[t]
    \centering
    \includegraphics[width=0.9\columnwidth]{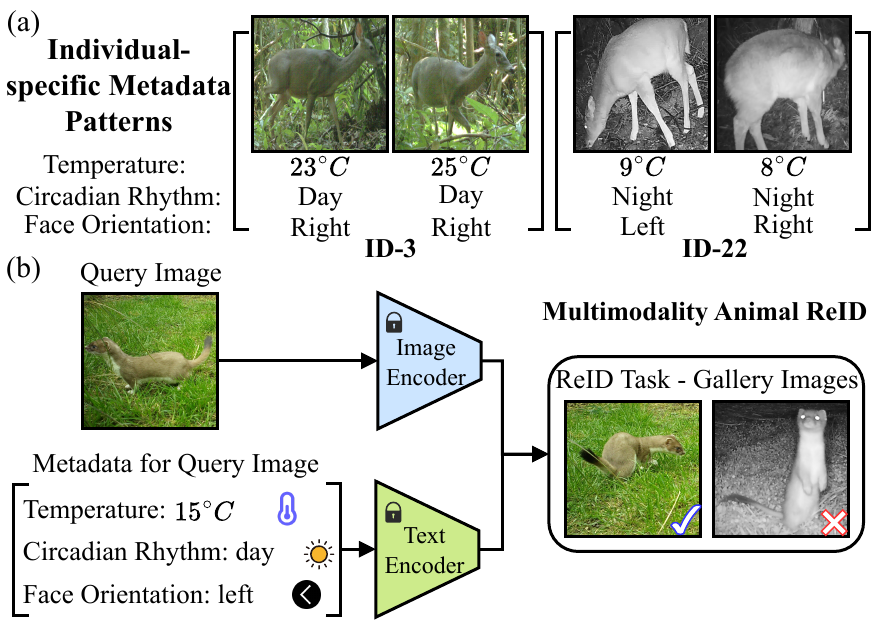}
    \caption{Overview of multimodal Animal ReID framework and the role of metadata. (a) Example images of two deer individuals showing distinct preferences across environmental conditions. (b) The model integrates visual features with textual environmental metadata.} 
    \label{fig:three_methods}
\end{figure}

Animal re-identification (Animal ReID) aims to recognize individual animals across images to support wildlife research such as population monitoring, movement ecology, and conservation management~\cite{schneider2019past,wu2026region,schofield2022more}. 
Compared to traditional tagging-based approaches, computer vision offers a scalable, non-invasive solution that avoids disturbing endangered species~\cite{beery2023wild,xu2024advanced}.
However, most existing Animal ReID datasets are purely visual, captured by surveillance or camera traps without contextual information~\cite{adam2024seaturtleid2022,li2020atrw,gao2021towards}.
This visual-only design introduces two limitations.
First, it omits environmental metadata (\textit{e.g.}, temperature and circadian rhythms), which ecologists have shown to be highly correlated with animal behavior and appearance~\cite{leliveld2022dairy}.
Figure~\ref{fig:three_methods}(a) shows that different individuals tend to appear under distinct environmental conditions, showing the potential of metadata to provide identity-discriminative cues.
Without such information, existing datasets cannot support the evaluation of the impact of incorporating environmental metadata on ReID performance.
Second, with the rapid advancement of multimodal models capable of jointly processing images and text~\cite{radford2021learning,qiao2026multiple,pang2025cabin}, image-only datasets underutilize their text-encoding capabilities, limiting the full potential of multimodal representations in Animal ReID.

To address these limitations, we present MetaWild, a dataset designed to enable systematic evaluation of metadata integration and multimodal learning in Animal ReID.
MetaWild is constructed from the publicly available New Zealand Trail Camera (NZ-TrailCams) dataset~\cite{nztrailcams}, a wildlife image collection captured using camera traps deployed in diverse natural habitats across New Zealand. 
We select six representative species, including invasive and native animals.
We have carefully curated images that are visually clear and contain reliable metadata overlays, comprising a total of 20,890 images. 
In addition to visual data, MetaWild provides curated environmental metadata, including \textit{temperature}, \textit{circadian rhythms}, and \textit{face orientation}, which are chosen based on their availability from overlays and their influence on animal appearance.
This design supports multimodal ReID by jointly exploiting visual and textual cues, as illustrated in Figure~\ref{fig:three_methods}(b).
Furthermore, since existing Animal ReID methods are generally designed to operate solely on visual inputs and cannot directly incorporate the textual metadata, we propose the Meta-Feature Adapter (MFA), a lightweight module that can be incorporated into existing VLM-based Animal ReID methods~\cite{jiao2024toward}, allowing the integration of environmental metadata into visual representations.

Our \textbf{contributions} are summarized as follows.
Firstly, we create and release MetaWild, a new dataset that pairs visual animal images with curated environmental metadata to evaluate the impact of incorporating environmental metadata on ReID performance, thereby facilitating the development of multimodal ReID methods.
Secondly, we propose MFA, a lightweight module that enables the integration of metadata into existing VLM-based Animal ReID models, allowing for performance evaluation with MetaWild without modifying the model architecture.
Finally, extensive experiments on MetaWild show that incorporating environmental metadata alongside visual data within a multimodal learning framework consistently improves Animal ReID performance.

\section{Related Work}
\noindent \textbf{Animal ReID} aims to recognize and match individual animals across images for wildlife monitoring~\cite{wu2026overcoming}. 
Existing methods include:
(1) \textit{Global feature learning} that treats the entire image as input, directly extracting global features for ReID~\cite{he2023animal}. 
(2) \textit{Species-specific feature extraction} that relies on distinctive local patterns (\textit{e.g.}, elephant ears~\cite{weideman2020extracting}), but they are sensitive to occlusions and viewpoint variations, which hinder their generalizability across species.
and (3) \textit{Auxiliary information integration}: Methods such as pose key point estimation~\cite{li2020atrw,li2025metawild} incorporate additional visual cues to refine feature extraction. 
Despite their success, these methods remain constrained to image-based data.

\noindent \textbf{Animal ReID Datasets.}
Existing datasets range from controlled farm or lab environments~\cite{kern2024towards,wang2026towermind,wahltinez2024open} to challenging in-the-wild camera-trap collections~\cite{li2020atrw}.
WildlifeDatasets~\cite{vcermak2024wildlifedatasets} provide a unified library and standardized tools for visual ReID benchmarking.
While these datasets have advanced visual-based ReID, they omit environmental metadata, which has been shown to influence animal behavior and appearance, limiting the evaluation of its impact on ReID performance.

\noindent \textbf{Vision-Language Model} learn aligned visual–textual representations~\cite{jia2021scaling,pang2025libra,zhao2025balancing,wang2026mars} and show strong generalization ability~\cite{zhao2024symmetric,zhao2026unlearning,qiao2026zero}.
While some recent works have adapted VLMs to ReID tasks~\cite{li2023clip,jiao2024toward}, they primarily leverage the image encoder and use the text encoder only for static, category-level descriptions.
In this paper, we leverage environmental metadata as a semantically rich textual information source to more fully utilize the text encoder for Animal ReID.

\section{The MetaWild Dataset}
\subsection{Dataset Composition}
MetaWild is designed to facilitate the evaluation of multimodal learning approaches in Animal ReID by pairing visual data with contextual environmental metadata. 
It is constructed from the publicly available NZ-TrailCams dataset~\cite{nztrailcams}.
To ensure the applicability and diversity of metadata integration across various animal species, MetaWild comprises 20,890 images spanning six representative species: Deer, Hare, Penguin, Pūkeko, Stoat, and Wallaby.
Each image is paired with environmental metadata extracted from embedded camera trap overlays.
\begin{figure}[t]
    \centering
    \includegraphics[width=0.9\linewidth]{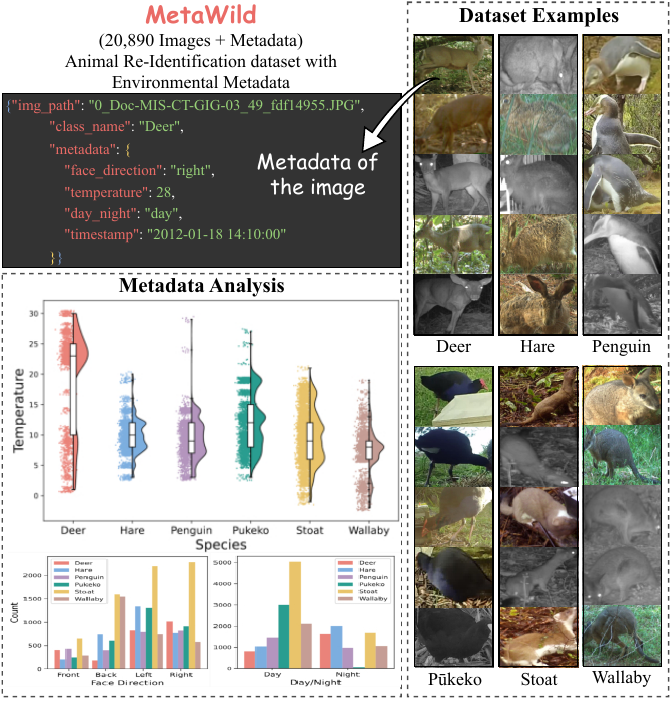}
    \caption{Examples of the MetaWild dataset.}
    \label{fig:dataset_illustration}
\end{figure}

\begin{table}[ht]
\setlength{\aboverulesep}{0.5pt}
\setlength{\belowrulesep}{0.5pt}
\centering
\setlength{\tabcolsep}{1.0mm}
\begin{tabular}{@{}l|rrrrrr|rr@{}}
\toprule
 & \multicolumn{2}{c}{Train} & \multicolumn{2}{c}{Gallery} & \multicolumn{2}{c|}{Query} & \multicolumn{2}{c}{Total} \\ \midrule
Datasets & Imgs & IDs & Imgs & IDs & Imgs & IDs & Imgs & IDs \\ \midrule
Deer & 1,631 & 21 & 586 & 17 & 216 & 17 & 2,433 & 38 \\
Hare & 1,820 & 31 & 926 & 29 & 306 & 29 & 3,052 & 60 \\
Penguin & 1,431 & 34 & 725 & 43 & 296 & 43 & 2,452 & 77 \\
Pūkeko & 1,854 & 11 & 800 & 19 & 411 & 19 & 3,065 & 30 \\
Stoat & 4,067 & 151 & 1,649 & 102 & 1,017 & 102 & 6,733 & 253 \\
Wallaby & 1,888 & 25 & 964 & 22 & 303 & 22 & 3,155 & 47 \\ \bottomrule
\end{tabular}
\caption{Details of Benchmark Datasets.}
\label{tab:benchmark_detail}
\end{table}

Figure~\ref{fig:dataset_illustration} presents examples of the species and environmental metadata included in the MetaWild dataset. 
The selection of species reflects conservation priorities, including predators (Stoat), pests (Wallaby, Hare), endangered native species (Yellow-eyed Penguin), and native animals (Deer and Pūkeko)~\cite{docpests,docnative}.
Stoats compete with native birdlife for food and habitat, also eat the eggs and young, and attack the adults, posing a significant threat to native wildlife. 
The Yellow-eyed Penguin, classified as endangered with a rapidly declining population, exemplifies a vulnerable native species in urgent need of protection. 
Wallabies and hares are considered agricultural pests, causing significant damage to native vegetation and ecosystems. 
Deer and Pūkeko, both native to New Zealand, exhibit unique visual and behavioral patterns that introduce diversity and realism, further enriching the dataset's applicability.
The dataset was organized according to standard ReID protocols, comprising 60\% for training, 25\% for the gallery, and 15\% for the query sets~\cite{li2018richly}, and detailed statistics are shown in Table~\ref{tab:benchmark_detail}.

\subsection{Dataset Construction}
We ensured high data quality by filtering visually clear images, performing reliable identity annotations, and extracting consistent metadata.
Both identity labels and metadata were independently verified by at least three annotators.
The detailed procedure is described below.

\noindent \textbf{Image Selection.}
We first conducted a filtering process to remove low-quality images in which the target animals appeared as unrecognizable, blurry blobs due to motion blur or poor lighting.
We further selected a representative subset of images for each species to ensure sufficient intra-species variation across individuals and conditions (\textit{e.g.}, viewpoints, lighting, and environments) and maintain balanced distribution across environmental metadata.

\noindent \textbf{Identity Annotation.}
We employed a combination of temporal analysis and visual verification to assign identities within each species.
For temporal analysis, we used the camera trap images' time-stamped nature to track animals across sequential frames.
Animals captured within narrow time windows (\textit{e.g.}, a few seconds) at the same camera location were likely to belong to the same individual.
To complement this, manual visual inspection was conducted to confirm or correct identity groupings based on distinct physical characteristics, such as markings, size, and shape.

\noindent \textbf{Metadata Extraction.}
We focus on three metadata features: \textit{temperature}, \textit{circadian rhythms}, and \textit{face orientation}, as they directly influence the animal's appearance and behavior~\cite{cade2021tools,mcvey2023invited}.
Temperature is read from embedded overlays, circadian rhythm (day/night) is inferred from timestamps and lighting, and face orientation is manually annotated to support geometric reasoning in ReID. Other metadata types were excluded due to redundancy or unavailability (\textit{e.g.}, missing geolocation). All metadata is standardized and stored in structured JSON files for each image.

\noindent \textbf{Image Preprocessing.}
To focus on the target animal and reduce background noise, we used a YOLO-based detector to generate bounding boxes for cropping, followed by manual verification to correct detection errors.
Furthermore, each cropped image was renamed using a structured format, $id\_camera$-$id\_count$ (\textit{e.g.}, $11\_CT$-$GIG$-$03\_27$, where $11$ denotes the individual identity, $CT$-$GIG$-$03$ represents the camera ID, and $27$ indicates the 28th image for identity 11).
This systematic naming convention facilitates efficient data management and traceability.

\begin{figure}[t!]
    \centering    \includegraphics[width=1.0\columnwidth]{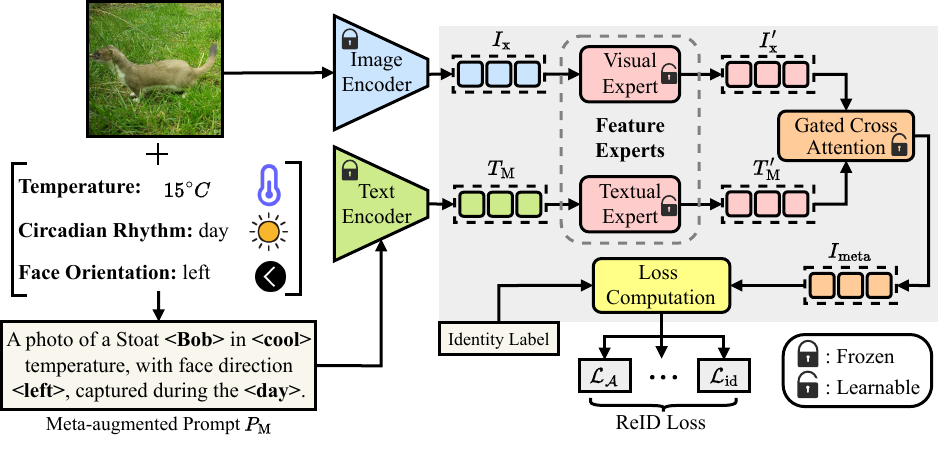}
    \caption{Overview of the proposed Meta-Feature Adapter (MFA) module}
    \label{fig:MFA_framework}
\end{figure}

\section{Methodology}
We propose a lightweight Meta-Feature Adapter (MFA) to integrate environmental metadata into existing VLM-based Animal ReID models without modifying their backbone architectures. 
As shown in Figure~\ref{fig:MFA_framework}, MFA consists of two main components:  
(1) \textbf{Feature Experts}, employed adapters~\cite{gao2024clip} as experts that refine visual and textual embeddings into metadata-aware representations.
(2) \textbf{Gated Cross-Attention}, which fuses visual and metadata features by weighting relevant information from each modality.

\subsection{Feature Experts}
To enable effective fusion between visual features and environmental metadata, we incorporate feature experts in both the text and image branches. 
In the text branch, we convert metadata into natural language descriptions using a fixed prompt template: ``A photo of a \{species\} \{individual id\} in \{freezing, cold, chilly, cool, warm, hot\} temperature, with face direction \{front, back, left, right\}, captured during the \{day, night\}." 
This template mimics natural language captions used during VLM pretraining, maximizing compatibility with the text encoder. 
The metadata-augmented prompt $P_{\text{M}}$ is encoded using the pretrained text encoder $\mathcal{T}(\cdot)$ to obtain text embedding $T_{\text{M}} = \mathcal{T}(P_\text{M})$, which is then refined by a Textual Metadata Expert (TME) $E_T$ into a metadata-aware embedding $T_\text{M}' = E_\text{T}(T_\text{M})$. 
Similarly, in the image branch, we introduce a Visual Feature Expert (VFE) $E_I$ to transform the raw visual embedding $I_\text{x}$ into metadata-aware representations $I_\text{x}' = E_I(I_\text{x})$. 
Both experts $E_T$ and $E_I$ are trained end-to-end using ReID loss functions including the identity classification loss $\mathcal{L}_{id}$ for encouraging separability across identities and the triplet loss $\mathcal{L}_{tri}$ for enforcing relative distance constraints~\cite{wang2021understanding}. 
By allowing gradients from the ReID objective to propagate through the feature experts, $E_T$ and $E_I$ learn task-relevant transformations that directly improve ReID performance.

\begin{table*}[t!]
\setlength{\aboverulesep}{0.5pt}
\setlength{\belowrulesep}{1.0pt}
\centering
\fontsize{9pt}{9pt}\selectfont
\setlength{\tabcolsep}{0.7mm}
\begin{tabular}{@{}l|cc|cc|cc|cc|cc|cc@{}}
\toprule
\multirow{2}{*}{Methods} & \multicolumn{2}{c|}{Deer} & \multicolumn{2}{c|}{Hare} & \multicolumn{2}{c|}{Penguin} & \multicolumn{2}{c|}{Pūkeko} & \multicolumn{2}{c|}{Stoat} & \multicolumn{2}{c}{Wallaby} \\ 
 & mAP & CMC-1 & mAP & CMC-1 & mAP & CMC-1 & mAP & CMC-1 & mAP & CMC-1 & mAP & CMC-1 \\ \midrule
CLIP-ZS~\cite{radford2021learning} & 50.0±.0 & 92.1±.0 & 33.8±.0 & 81.1±.0 & 34.0±.0 & 57.8±.0 & 33.0±.0 & 69.6±.0 & 30.1±.0 & 72.7±.0 & 46.2±.0 & 85.5±.0 \\ \midrule
CLIP-FT & 63.2±.1 & 95.4±.4 & 56.7±.2 & 92.5±.4 & 44.0±.3 & 64.9±.2 & 56.8±.1 & 80.3±.5 & 68.6±.1 & 92.3±.3 & 55.5±.1 & 92.4±.4 \\
CLIP-FT+MFA & \textbf{66.7±.2} & \textbf{95.7±.3} & \textbf{58.4±.3} & 92.6±.2 & \textbf{46.0±.2} & 64.9±.3 & \textbf{58.2±.2} & \textbf{81.6±.3} & \textbf{69.8±.2} & 91.8±.2 & \textbf{56.8±.4} & 90.7±.3 \\ \midrule
CLIP-ReID~\cite{li2023clip} & 65.2±.4 & 95.8±.3 & 60.0±.6 & 95.1±.3 & 44.8±.4 & 67.9±.4 & 57.6±.2 & 82.0±.1 & 67.5±.1 & 91.5±.3 & 56.9±.4 & 88.8±.2 \\
CLIP-ReID+MFA & \textbf{69.4±.2} & \textbf{98.1±.1} & \textbf{63.2±.1} & 95.4±.1 & \textbf{50.3±.4} & \textbf{68.6±.2} & \textbf{59.8±.1} & \textbf{83.7±.2} & \textbf{71.5±.2} & \textbf{92.0±.1} & \textbf{61.8±.2} & \textbf{92.1±.1} \\ \midrule
ReID-AW~\cite{jiao2024toward} & \multicolumn{1}{l}{67.5±.3} & 96.0±.2 & 63.3±.4 & 95.6±.3 & 48.8±.5 & 69.4±.3 & 58.5±.3 & 82.0±.4 & 69.5±.3 & 93.5±.5 & 58.4±.3 & 91.8±.2 \\
ReID-AW+MFA & \textbf{72.4±.2} & \textbf{97.0±.2} & \textbf{66.2±.3} & \textbf{96.8±.2} & \textbf{55.3±.4} & \textbf{70.8±.4} & \textbf{61.8±.2} & \textbf{86.7±.3} & \textbf{74.1±.4} & \textbf{95.0±.2} & \textbf{63.5±.1} & \textbf{92.7±.2} \\ \bottomrule
\end{tabular}
\caption{Intra-species re-identification performance on the MetaWild dataset across six species, we report mAP and CMC-1 accuracy (\%) with 95\% confidence intervals. CLIP-ZS shows zero variance due to its deterministic zero-shot inference nature.}
\label{tab:evaluation_results_intra}
\end{table*}

\begin{table*}[t!]
\setlength{\aboverulesep}{0.5pt}
\setlength{\belowrulesep}{1.0pt}
\centering
\fontsize{9pt}{9pt}\selectfont
\setlength{\tabcolsep}{0.7mm}
\begin{tabular}{@{}l|cc|cc|cc|cc|cc|cc@{}}
\toprule
\multirow{2}{*}{Methods} & \multicolumn{2}{c|}{Deer} & \multicolumn{2}{c|}{Hare} & \multicolumn{2}{c|}{Penguin} & \multicolumn{2}{c|}{Pūkeko} & \multicolumn{2}{c|}{Stoat} & \multicolumn{2}{c}{Wallaby} \\ 
 & mAP & CMC-1 & mAP & CMC-1 & mAP & CMC-1 & mAP & CMC-1 & mAP & CMC-1 & mAP & CMC-1 \\ \midrule
CLIP-ZS~\cite{radford2021learning} & 40.3±.0 & 80.5±.0 & 25.1±.0 & 75.1±.0 & 27.5±.0 & 50.4±.0 & 24.6±.0 & 64.5±.0 & 23.0±.0 & 62.3±.0 & 40.6±.0 & 80.3±.0 \\ \midrule
CLIP-FT & 55.1±.2 & 83.0±.4 & 41.5±.2 & 81.4±.2 & 38.7±.3 & 59.7±.2 & 41.3±.2 & 76.9±.2 & 45.4±.3 & 79.3±.2 & 49.0±.2 & 72.2±.1 \\
CLIP-FT+MFA & \textbf{56.7±.2} & \textbf{86.4±.4} & \textbf{43.9±.4} & 81.6±.2 & \textbf{40.8±.1} & \textbf{63.5±.2} & \textbf{42.3±.3} & \textbf{76.9±.4} & \textbf{46.2±.4} & 79.6±.2 & \textbf{50.0±.3} & \textbf{72.8±.3} \\ \midrule
CLIP-ReID~\cite{li2023clip} & 56.3±.3 & 84.4±.3 & 43.5±.1 & 86.3±.3 & 39.5±.2 & 60.3±.4 & 43.8±.4 & 77.9±.1 & 45.8±.3 & 79.3±.2 & 50.1±.4 & 82.5±.2 \\
CLIP-ReID+MFA & \textbf{60.2±.4} & \textbf{89.2±.3} & \textbf{44.1±.1} & \textbf{88.6±.2} & \textbf{41.3±.2} & \textbf{64.1±.4} & \textbf{45.1±.3} & \textbf{78.3±.2} & \textbf{47.9±.1} & \textbf{80.3±.2} & \textbf{52.3±.2} & \textbf{82.8±.2} \\ \midrule
ReID-AW~\cite{jiao2024toward} & 59.3±.3 & 89.0±.2 & 47.6±.2 & 90.2±.4 & 40.8±.5 & 63.9±.3 & 50.4±.4 & 80.3±.1 & 53.3±.3 & 83.9±.1 & 51.7±.3 & 84.2±.2 \\
ReID-AW+MFA & \textbf{62.5±.4} & \textbf{92.4±.4} & \textbf{50.2±.3} & \textbf{90.8±.2} & \textbf{44.2±.4} & \textbf{64.6±.3} & \textbf{53.6±.3} & \textbf{83.6±.1} & \textbf{56.1±.1} & \textbf{84.6±.1} & \textbf{53.1±.2} & \textbf{86.0±.3} \\ \bottomrule
\end{tabular}
\caption{Leave-one-domain-out inter-species ReID performance on the MetaWild dataset, we report mAP and CMC-1 accuracy (\%) with 95\% confidence intervals for each target species.}
\label{tab:evaluation_results_cross}
\end{table*}

\subsection{Gated Cross-Attention}
We utilize a cross-attention mechanism to integrate metadata-aware text embeddings $T'_{\text{M}}$ with visual features $I'_{\text{x}}$. 
Unlike conventional fusion methods, cross-attention enables context-aware integration, allowing image features to selectively attend to relevant metadata cues. 
Given image embeddings $I_\text{x}' \in \mathbb{R}^{N \times d}$ and metadata-aware text embeddings $T_\text{M}' \in \mathbb{R}^{M \times d}$, we compute Query ($Q$), Key ($K$), and Value ($V$) matrices as $Q = I_\text{x}' W_Q$, $K = T_\text{M}' W_K$, $V = T_\text{M}' W_V$, where $W_Q$, $W_K$, and $W_V$ are learnable projection matrices. 
To address the fact that metadata relevance varies across images, we employ a gating mechanism to selectively adjust metadata contribution. 
A gating value $\gamma \in [0, 1]$ is computed as $\gamma = \text{Gate}(I_\text{x}', T_\text{M}') = \sigma(\text{MLP}([I_\text{x}'; T_\text{M}']))$, where $[I_\text{x}'; T_\text{M}']$ denotes concatenation, MLP is a multi-layer perceptron with layer normalization, and $\sigma$ is the sigmoid activation. 
The final meta-augmented image embedding $I_{\text{meta}}$ is computed as $I_{\text{meta}} = \gamma AV + I_\text{x}'$, where $A$ is the cross-attention weight matrix~\cite{shi2022dense}.
The loss function is defined as:
\begin{equation}
    \label{eq:loss_attention}
    \mathcal{L}_{\mathcal{A}}^i = -\log \frac{\exp\left( s\left( T_{\text{M}}'^{i}, I_{\text{meta}}^{i} \right) / \tau \right)}{\sum_{j=1}^{B} \exp\left( s\left( T_{\text{M}}'^i, I_{\text{meta}}^{j} \right) / \tau \right)}
\end{equation}
where $B$ is the batch size, $( T_{M}'^i, I_{\text{meta}}^i)$ is the $i\text{-th}$ matched pair, and $\tau$ is the temperature parameter.

\section{Experiments}
We evaluate existing Animal ReID models under visual-only and visual+metadata settings on MetaWild using two protocols: 
(1) \textbf{Intra-species ReID}~\cite{varghese2023fine}, where training and testing are performed on different individuals within the same species; and 
(2) \textbf{Inter-species ReID}~\cite{heiling2016using}, where we adopt a leave-one-domain-out (LODO) strategy~\cite{yu2024rethinking}, training on five species and testing on the remaining unseen species to reflect real-world scenarios where collecting labeled data for every species is impractical~\cite{jiao2024toward}.

\subsection{Experimental Results}
\noindent \textbf{Intra-species ReID.}
Table~\ref{tab:evaluation_results_intra} shows that incorporating environmental metadata consistently improves ReID performance across all six species. 
CLIP-ReID achieves mAP gains of 5.5\% on Penguin, 4.9\% on Wallaby, and 4.2\% on Deer, while ReID-AW shows improvements of 6.5\% on Penguin, 5.1\% on Wallaby, and 4.9\% on Deer, demonstrating the effectiveness of metadata integration.

\noindent \textbf{Inter-species ReID.}
Table~\ref{tab:evaluation_results_cross} summarizes LODO evaluations where one species is held out for testing while the remaining five are used for training.
Incorporating metadata consistently improves all baseline methods: CLIP-ReID+MFA achieves mAP gains of 3.9\% on Deer, 3.3\% on Pūkeko, and 1.8\% on Penguin, while ReID-AW shows improvements of 3.4\% on Penguin, 3.2\% on Deer, and 2.6\% on Hare. 
These results demonstrate that environmental metadata provides complementary cues that enhance model generalization across species boundaries and improve transferable identity representation learning.

\section{Conclusion}
We present MetaWild, a multimodal Animal ReID dataset pairing visual data with environmental metadata to evaluate metadata's impact on ReID performance.
To support this investigation without architectural changes, we further propose the Meta-Feature Adapter (MFA), a lightweight module that enables the integration of metadata into VLM-based Animal ReID methods. 
Extensive experiments demonstrate that incorporating metadata alongside visual information consistently improves ReID accuracy, confirming the value of contextual environmental metadata. 
We hope this work can inspire broader exploration of environmental metadata and multimodal approaches in wildlife ReID and beyond.

\bibliography{main}


\end{document}